\documentclass[final,5p,times,twocolumn]{elsarticle}

\usepackage{amssymb}

\usepackage{lineno}
\usepackage{times}
\usepackage{epsfig}
\usepackage{graphicx}
\usepackage{amsmath}
\usepackage{amssymb}

\usepackage{url}
\usepackage{times}
\usepackage{epsfig}
\usepackage{graphicx}
\usepackage{amsmath}
\usepackage{amssymb}
\usepackage{comment}
\usepackage{subfig}
\usepackage{hhline} 
\usepackage{float}
\usepackage{mathtools}
\usepackage{dblfloatfix}

\newcommand\independent{\protect\mathpalette{\protect\independenT}{\perp}}
\def\independenT#1#2{\mathrel{\rlap{$#1#2$}\mkern2mu{#1#2}}}

\journal{}

\begin{document}

\begin{frontmatter}



\title{Investigating Bias in Deep Face Analysis: The KANFace Dataset and Empirical Study}

\author[label1]{Markos Georgopoulos}
\author[label2]{Yannis Panagakis}
\author[label1]{Maja Pantic}
\address[label1]{Department of Computing, Imperial College London, UK}
\address[label2]{Department of Informatics and Telecommunications, University of Athens, Greece}

\author{}

\address{}

\begin{abstract}
Deep learning-based methods have pushed the limits of the state-of-the-art in face analysis. However, despite their success, these models have raised concerns regarding their bias towards certain demographics. This bias is inflicted both by limited diversity across demographics in the training set, as well as the design of the algorithms.
In this work, we investigate the demographic bias of deep learning models in face recognition, age estimation, gender recognition and kinship verification.
To this end, we introduce the most comprehensive, large-scale dataset of facial images and videos to date. It consists of $40$K still images and $44$K sequences ($14.5$M video frames in total) captured in unconstrained, real-world conditions from $1,045$ subjects. The data are manually annotated in terms of identity, exact age, gender and kinship. The performance of state-of-the-art models is scrutinized and demographic bias is exposed by conducting a series of experiments.
Lastly, a method to debias network embeddings is introduced and tested on the proposed benchmarks. 
\end{abstract}


\end{frontmatter}

\section{Introduction}\label{sec:introduction}

Computational models of human face relying on deep neural networks have significantly advanced the state-of-the-art in face analysis \cite{masi2018deep}, \cite{Ranjan:Hyperface} and photorealistic face generation \cite{vougioukas2019realistic, siarohin2019first}, among many other computer vision tasks \cite{liu2020deep, minaee2020image}. These advances would not have been made possible without the continuous efforts of the research community in collecting and annotating diverse datasets of human faces.
At the same time, models of higher expressivity and learning capacity have been developed in order to handle transformations that are present in these datasets.
In this spirit, early facial datasets that were captured in controlled environments (e.g., \cite{pie}) are now being replaced by extensive datasets of faces captured "in-the-wild" (e.g., \cite{LFW,Megaface}), while simple linear models (e.g., \cite{Turk:Eigenfaces}) are being replaced by deep learning algorithms (see \cite{masi2018deep} for a survey of such methods).

Despite these efforts, capturing the complex variability in facial data is an almost intractable task, given a limited amount of resources. Hence, any finite set of facial data describes only certain aspects of facial diversity, resulting in \textit{dataset bias} with regards to an underrepresented instance of the real world \cite{torralba:dataset}. 
On the other hand, designing and training a model that is invariant to all possible face variability factors is also rather intractable. Consequently, any face analysis model suffers from \textit{algorithmic bias}, i.e., it cannot generalize on sources of variation that are not explicitly modeled and are underrepresented in the training set.
As of late, algorithmic discrimination has become an issue, with commercial systems reportedly demonstrating bias with regards to gender and skin color \cite{gender_shades}.  
In turn, and as AI became indispensable in everyday's decision making, research on the risks and mitigation of bias in AI has attracted both commercial and academic interest.

Dataset bias comes in many forms and can be categorized based on the aspect of the visual world that is absent from the data distribution. A usual source of bias in early datasets is the lack of diversity in capturing conditions, e.g., environment, recording device and head pose. For example, PIE \cite{pie} contains faces in controlled poses that were captured in lab environment. Another main source of bias lies in the demographics of the recorded people, in other words, the different semantic categories (e.g., gender, age group and skin color).  
For instance, one third of the images in LFW \cite{LFW} contain faces of people who are over 60 years old, while CELEB-A\cite{Celeba} contains mainly faces with lighter skin.
Furthermore, the collection of large-scale datasets (e.g., \cite{Cao:VGGface2,Rothe:IJCV16}) usually involves downloading images from the web using semi-automatic pipelines, a procedure that can also introduce unwanted bias. In particular, scraping websites such as Flickr and Google for images can lead to datasets that inherit the bias of the source (e.g., the search engine) as well as the cultural bias \cite{ownage_bias,owngender_bias,ownrace_bias} of the data collector (e.g., the input query) \cite{Zou:Nature}.
The bias can also be specific to the task for which the dataset is collected. For instance, cropping face pairs from the same image can affect the task of kinship verification significantly by adding factors like the environment, chrominance and image quality \cite{Lopez:TPAMI16,Georgopoulos:IMAVIS18}.

Overall, dataset bias is a critical subject in computer vision that has been studied thoroughly \cite{torralba:dataset,tommasi2017deeper} due to its impact on the performance of trained models. However, algorithmic bias does not originate solely from the demographic disparities in the training set. 
Immaculate modeling of the complex variability and transformations of the human face is still intractable.
Therefore, the assumptions made by the model, i.e., its inductive bias, will not always hold. For example, methods that rely on least squares error minimization fail to account for sparse gross errors in the training data. Hence, such models are not able to handle transformations like eye-glasses and occlusions, that occur naturally on facial images in-the-wild.

In this work, we investigate demographic bias in deep face analysis. To this end, we introduce KANFace dataset (Kinship, Age, geNder): the largest manually annotated image and video dataset. The diversity of the faces in the dataset is quantitatively evaluated and demonstrated in baseline experiments. We focus on age and gender bias and perform experiments on face recognition, age estimation, gender recognition and kinship verification. By leveraging the abundance of annotations, the performance of the baseline models is diagnosed and the biased behavior of the state-of-the-art baselines is exposed. Lastly, in order to alleviate the demographic disparities in model performance, we introduce a method to debias the pretrained network embeddings. To summarize:

\begin{itemize}
    \item We present the KANFace dataset- the most comprehensive, manually collected dataset of facial images and videos consisting of $41,036$ images and $14.5M$ video frames, captured in unconstrained real-world conditions from $1,045$ subjects. Specifically, for each subject, our dataset includes $39$ images and $13,870$ frames on average, captured across different ages (from $0$ to $100$ years). The dataset is annotated with regards to identity, age, gender and kinship. The age annotations make the proposed dataset the first aging video dataset captured in-the-wild, as it contains video sequences of each subject at multiple time instances. 
    Moreover, more than half of the subjects in the dataset are related to each other. With 566,198 kin pairs, this is not only one of the largest datasets with kinship annotations but also the first large-scale dataset to allow for video-based kinship verification and recognition in-the-wild. 
    A comparison to other publicly available face datasets is given in Table \ref{tab:datasets}. The characteristics of the dataset and the collection process are described in detail in Section \ref{s:dataset}.

    \item We quantitatively assess the diversity of faces in the proposed dataset in terms of age, gender, craniofacial ratios, facial region contrast and skin-color tone by adopting the scheme proposed in \cite{DiF}. In addition, we quantify the impact of facial diversity with baseline experiments in Section \ref{s:baseline_experiments}. The results highlight the challenges posed by our dataset compared to other standard benchmarks.

    \item The bias of deep learning systems across sensitive demographics, namely across ages and gender as well as perceived skin-color tone is investigated in Section \ref{s:bias_analysis}. In particular, using our benchmark, we evaluate the bias of different deep architectures trained on different datasets. Lightweight models are also considered due to their wide adoption in the industry. Moreover, we study the generalization ability of deep age and gender recognition methods by conducting comprehensive cross-dataset experiments. This is the first study of its kind, the results of which shed light on the impact of training data and choice of architecture on the performance of a model.
    
    \item Lastly, we investigate the mitigation of unwanted bias in pretrained network embeddings in Section \ref{s:debias}. In particular, we study whether certain demographic attributes can be disentangled from the representation without constraining its discriminative ability. To this end, we propose a framework that decomposes each feature into the sum of a task-specific term and the unwanted bias terms. The method is used to debias the embeddings obtained from the baseline models for face recognition, age estimation and gender recognition. The experimental results uncover the redundancy of deep embeddings, reveal the correlation between identity, age and gender, and indicate in which cases theses attributes can be disentangled.
\end{itemize}

\begin{table*}[hbt!]
\centering  
\resizebox{\linewidth}{!}{
\begin{tabular}{llllllll}

\hline
Dataset&\#Images &\# Sequences&\# Frames&\# Identities& age-range&\#kin pairs&Labels\\
\hline
FG-NET \cite{FGNET},\cite{Lanitis:TPAMI02prog}& 1,002 &-&-& 82 & 0-69  & - & ID, EA    \\
MORPH 2 \cite{Morph}& 55,134&-&- & 13,618 & 16-77 & -&ID, EA, G  \\
LFW \cite{LFW}& 13,233 &-&-& 5,749 & - & - & ID   \\
CornellKin \cite{Fang:ICIP10}& 300   &-&- &-&- & 150 & K  \\
YTF \cite{YTF}&-&3,425&62,095&1,595&-&-& ID\\
UBKinface \cite{Xia:IJCAI11}, \cite{Shao:CVPRw11}& 600   &  - & -     & -&-&400   & K    \\
UvA-NEMO \cite{Dibekliouglu:ECCV12}&-&1240&N/A&400&8-76&101&ID, EA, K  \\
OUI-Adience \cite{Eidinger:TIFS14}& 26,580&-&- & 2,284 & 0-60+ & - & ID, AG, G   \\
CACD \cite{Chen:TransMult15}& 163,446 &-&-& 2,000 & N/A& - & ID, EA    \\
CASIA-Webface \cite{casia_webface}& 494,414 &-&-& 10,575 & - & - & ID   \\
KinFaceW-I \cite{Lu:TPAMI14}& 1,066 &    -& -&-&   -  & 533   & K  \\
KinFaceW-II \cite{Lu:TPAMI14}& 2,000 &  -& -&-& -  & 1,000   & K   \\

VGGFace \cite{Parkhi:BMVC15}&2.6M&-&-&2,622&-&-&ID\\
UMDFaces \cite{UMDfaces},\cite{Bansal:ICCVW17}&367,888&22,075&3.7M&8,277/3,107&-&-&ID, G\\
IMDB-WIKI \cite{Rothe:IJCV16}& 523,051 &-&-& 20,284 & 0-100 & - & ID, EA, G   \\
MS-Celeb-1M \cite{msceleb1m}& 10M &-&-& 100,000 & - & - & ID   \\
FIW \cite{FIW}, \cite{robinson2018visual} &   30,725 &-&-& 10,676 &   -   &  656,954  &   ID, G, K \\
MegaFace \cite{Megaface}& 4.7M &-&-& 690,572 & - & - & ID  \\
IJB-B \cite{IJB-B} &21,798&7,011&55,026&1,845&-&-&ID\\
VGGFace2 \cite{Cao:VGGface2}&3.31M &-&-&9,131&-&-&ID,AG, G\\
\hline
\textbf{KANFace}  &41,036&\textbf{44,224}& \textbf{14.5M}& 1,045 &0-100&566,198&\textbf{ID, EA, G, K}\\

\hline

\end{tabular}
}
\caption{Comparison of image and video datasets for facial modeling. ID:identity, EA: exact age, AG:age group, G:gender, K:kinship}
\label{tab:datasets}
\end{table*}

\section{The proposed dataset}\label{s:dataset}

To ensure the absence of noisy labels, KANFace dataset was manually collected and annotated. This is a particularly laborious process for such a large-scale dataset. In this section, we introduce the collection and annotation strategy. 

\begin{figure}[h!]
  \centering

    \includegraphics[width=0.35\textwidth]{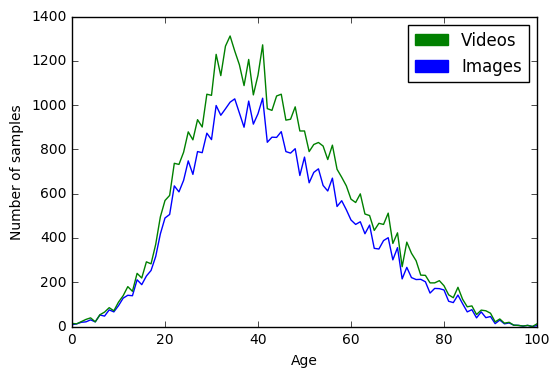}

  \caption{The age distribution of the dataset}
 
  \label{fig:ages}
\end{figure}

\textbf{Creating the list of subjects:} Since our main goal is the collection of a rich video dataset, we focus on public figures, e.g., actors, musicians and politicians, who have an abundance of available videos online from various periods of their lives. Our search is constrained by gender balance and the existence of family connection between the subjects. By capitalizing on the available metadata on the \textit{Wikipedia} and \textit{IMDb} websites, we deliver a list of $1,045$ ($586$ male and $459$ female) identities and $544$ kin related subject pairs. The age distribution of the dataset is presented on Figure \ref{fig:ages}.

\textbf{Manual collection of the videos: } We only focus on YouTube videos, the "capture date" of which was either annotated by the video uploader or could be acquired from the \textit{IMDb} or \textit{Wikipedia} page of the celebrity. The age of the subject is calculated by subtracting their date of birth from the capture date and is therefore accurate to the year. The final video dataset consists of $13$ videos per subject on average, all of which were captured at different ages. The context of these videos varies greatly and includes interviews, movies, sport events and speeches, to name but a few.

\textbf{Final sequence extraction: } Only the frames that contain the subject of interest (SOI) were manually located, annotated and kept.

\textbf{Face detection and subject recognition: }The proposed dataset contains bounding box annotations for each video frame, which were obtained as follows : 1) A face detector \cite{mtcnn} was employed in order to detect all the faces in each frame. 2) A maximum of $20$ frames were then uniformly sampled from each sequence. These frames were selected to act as anchor frames. 3) A gallery dataset, consisting of $10$ images per subject, is collected for each subject from \textit{Google Images}. 4) The SOI in each anchor frame was recognized by comparing the deep embeddings \cite{facenet} of the faces in the frame with the gallery dataset. 5) The selected face in each anchor frame was manually verified or corrected.

\textbf{Bounding box selection: }The manually verified face boxes of the anchor frames were used to select the face bounding boxes of the intermediate frames. In particular, we base our strategy on the assumption that the position of the face bounding box does not change significantly between consecutive frames. To quantify the movement of the face box, we utilized the intersection over union (IOU) metric. Concretely, starting with the face box of an anchor frame, the IOU with every detected face on the next and previous frames was calculated. The bounding boxes with the highest score were selected and the process continued bidirectionally.

\begin{figure}[h!]
  \centering

    \includegraphics[width=0.35\textwidth]{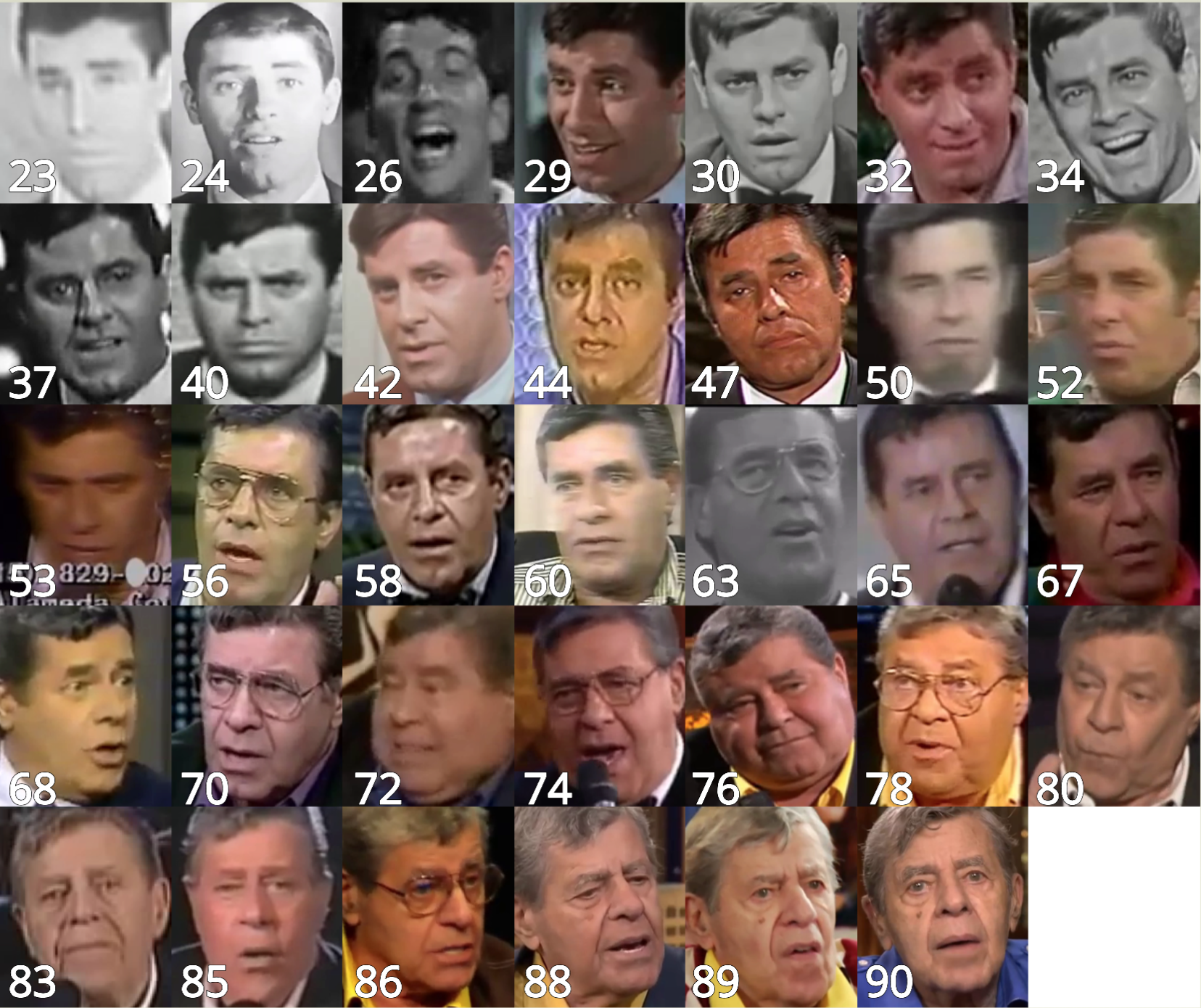}

  \caption{Images from the proposed dataset depicting the same person at 35 different ages.}
 
  \label{fig:ageprog}
\end{figure}

\begin{figure*}[t!]
  \centering

    \includegraphics[width=415pt]{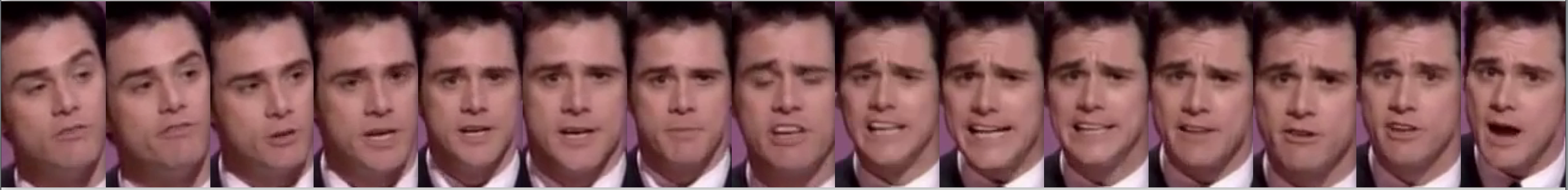}
    \includegraphics[width=415pt]{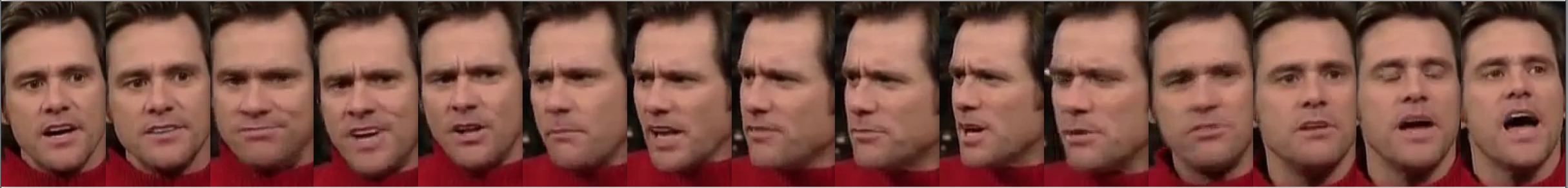}
    \includegraphics[width=415pt]{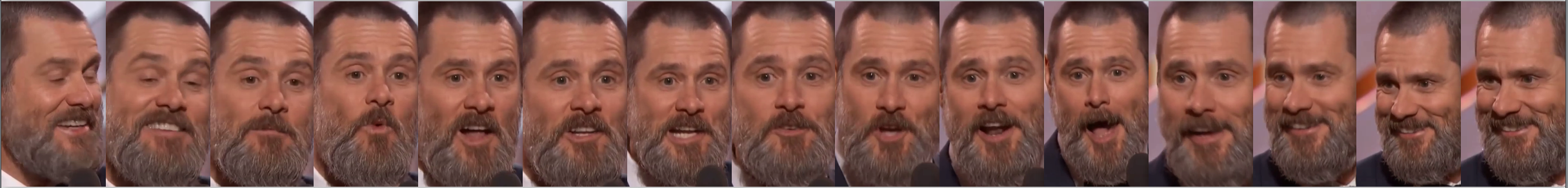}
  
  \caption{Sample sequences of the same person at 3 different ages at a 5 frame interval.}
 
  \label{fig:videos}
\end{figure*}

\textbf{Manual refinement and selection of images: }In the final stage of the pipeline the selected face bounding boxes in all frames of the dataset are manually verified. At the same time, a small number of images per sequence were selected to form the static version of the dataset. These images were manually selected, in order to avoid frames with motion blurriness. The selected images display large variation with regards to pose, expression, illumination, occlusions and image quality. Sample images and sequences from our dataset are depicted in Figures \ref{fig:ageprog} and \ref{fig:videos}.

\textbf{Considerations regarding the data collection: }
The nature of the proposed dataset posed several challenges with regards to its collection. In particular, as an aging dataset, KANFace contains videos of people at multiple ages, that typically span decades. At the same time, we wanted a large portion of the subjects to be related. The above reasons rendered a human study impossible. Therefore, we decided to use publicly available YouTube videos of celebrities for our research.

\section{Diversity of the dataset}
\subsection{Diversity metrics}\label{div_analysis}

In order to quantify the diversity of the faces in our dataset, we follow the scheme introduced in \cite{DiF}. In particular, we evaluate the diversity with regards to the labels of the data (e.g., age, gender and perceived skin color) and capturing conditions (e.g. pose, image quality and illumination). Concretely, we measure statistics for: age, gender, pose, inter-ocular distance, resolution, craniofacial ratios (i.e., CF0-CF9), facial region contrast (i.e., FC0-FC8), skin-color tone and illumination (as measured by individual typology angle).

The diversity of the dataset is quantified using the Simpson and Shannon diversity indices, which are borrowed from biodiversity studies \cite{heip1998indices}. Simpson D and Shannon H measure the diversity of the dataset, while Simpson E and Shannon E quantify the evenness of the distribution. The diversity and evenness indeces are calculated as follows:
\begin{alignat}{3}
    &Shannon: \; \; &&H = - \sum_1^Sp_i\; ln(p_i), \quad &&E = \frac{H}{ln(S)} \nonumber\\
    &Simpson: &&D = \frac{1}{\sum_1^Sp_i^2} , &&E = \frac{D}{S} \nonumber
\end{alignat}
where $p_i$ is the probability of class $i$ and $S$ is the number of classes. The mean, standard deviation, Simpson D (SiD) and E (SiE) and Shannon H (ShH) and E (ShE) for the attributes of the dataset are tabulated in Table \ref{tab:diversity}. The diversity statistics are similar to \cite{DiF}, particularly with regards to resolution, craniofacial ratios and facial contrast. Furthermore, the proposed dataset is significantly more diverse with regards to age labels, which were, contrary to \cite{DiF}, manually annotated.

\begin{table}[hbt!]
\centering    
\begin{tabular}{|c|c|c|c|c|c|}

\hline
\textbf{Att.}&\textbf{SiD}&\textbf{SiE}&\textbf{ShH}&\textbf{ShE}&\textbf{Mean (Std)}\\
\hline
 Age&59.58&0.59&4.221&0.915&43.13(16.98)\\
\hline
 Gender&1.999&0.999&0.999&0.999&0.51(0.5)\\  
\hline
 Pose&2.42&0.303&1.107&0.533&1.013(1.25)\\
\hline
 IOD&5.55&0.694&1.851&0.89&55.2(32.57)\\
 Res.&3.921&0.56&1.584&0.814&128(72.13)\\
\hline
 CF0&5.8&0.967&1.775&0.99&0.875(0.065)\\
 CF1&5.8&0.967&1.775&0.99&0.5276(0.062)\\
 CF2&5.867&0.978&1.78&0.994&0.442(0.019)\\
 CF3&5.895&0.982&1.783&0.995&0.63(0.058)\\
 CF4&5.889&0.981&1.782&0.995&0.64(0.062)\\
 CF5&5.75&0.958&1.77&0.988&0.461(0.06)\\
 CF6&5.51&0.918&1.747&0.975&0.467(0.058)\\
 CF7&5.86&0.978&1.78&0.994&0.541(0.054)\\
 CF8&5.693&0.949&1.765&0.985&0.399(0.16)\\
 CF9&5.812&0.969&1.775&0.99&0.378(0.037)\\
 \hline
 FC0&5.893&0.929&1.782&0.995&0.772(0.071)\\
 FC1&5.572&0.929&1.757&0.981&0.722(0.051)\\
 FC2&5.592&0.932&1.758&0.981&0.723(0.051)\\
 FC3&5.854&0.976&1.779&0.993&0.3(0.107)\\
 FC4&5.588&0.93&1.755&0.98&0.203(0.074)\\
 FC5&5.6686&0.945&1.763&0.984&0.215(0.074)\\
 FC6&5.919&0.985&1.785&0.996&0.72(0.083)\\
 FC7&5.837&0.973&1.778&0.992&0.675(0.075)\\
 FC8&5.835&0.972&1.778&0.992&0.675(0.075)\\
 \hline
 ITA&3.239&0.54&1.365&0.762&32.72(9.39)\\

\hline

\end{tabular}

\caption{Diversity analysis of the dataset. IOD: inter-ocular distance, res: resolution, CF0: facial index, CF1: mandibular index, CF2: intercanthal index, CF3: left orbital width index, - CF4: right orbital width index, CF5: left eye fissure index, CF6: right eye fissure index, CF7: nasal index, CF8: vermilion height index, CF9: mouth-face width index, FC0/FC1/FC2: eyes region contrast, FC3/FC4/FC5: lips region contrast, FC6/FC7/FC8: eyebrows region contrast and ITA: individual typology angle. For an analysis of these indeces, the reader is referred to \cite{DiF}.}
\label{tab:diversity}
\end{table}

\subsection{Baseline experiments}\label{s:baseline_experiments}

To further assess the diversity of the faces in KANFace, we investigate its impact on model performance. We evaluate a baseline model against different benchmarks for face analysis tasks, namely face recognition, age estimation, gender recognition and kinship verification. A diverse dataset is bound to be more challenging due the presence of multiple modes of variation. Hence, we expect a significant performance drop on the KANFace image and video benchmarks.

Due to their efficacy in numerous computer vision tasks (e.g. object detection \cite{rcnn} and recognition \cite{resnet}), we choose a deep CNN to obtain the facial representation for our baseline model. In particular, we design our network based on the VGG-16 \cite{Simonyan:VGG16} architecture. 
For each of the studied tasks, the model is trained on a different dataset and fine-tuned on KANFace.
Similarly, to extract a temporal representation from the video sequences, we utilize Recurrent Neural Network (RNN) on the CNN embedding of each frame. For the experiments on videos, the CNN features are obtained from the output of the first fully connected layer and are fed to a gated feedback recurrent neural network (GRU) \cite{GRU}. 
The final video representation is obtained from the $512$-dimensional hidden state of the last frame of the sequence.

\subsubsection{Baseline experiments on face recognition}\label{sub:base_face_reco}

Our face recognition model is trained on the extensive VGGFace dataset. The resulting 4,096-dimensional face descriptor is efficiently discriminative and achieves competitive performance on standard face recognition benchmarks (e.g. 98.95\% on LFW). In the case of videos, the GRU is trained on the proposed video dataset in a many-to-one configuration, as described in the previous section. The static and temporal embeddings are classified through a softmax classification layer with 1,045 neurons.

\textbf{Evaluation protocol and baseline results:}
The proposed dataset contains 39 images and 42 sequences on average per person. The number of data per subject raises the issue of limited intraclass variability between different images or sequences of a person at the same age. In such cases, the data may originate from the same video and have similar attributes (e.g. environment, image quality, chrominance). In order to deal with the related bias, each combination of identity and age is only included in either the training or test set. Consequently, our baseline experiments follow the paradigm of age-invariant face recognition.
We perform 5-fold cross-validation and report the mean classification accuracy and standard deviation in Table \ref{tab:face_reco_im}. The results are compared to those obtained for the LFW, YTF, FG-NET and CACD datasets. In the absence of a face recognition benchmark for these aging datasets, a random 80-20 split protocol is used for evaluation.

\begin{table}[hbt!]
\centering    
\begin{tabular}{|c|c|c|}

\hline
\textbf{Model}&\textbf{Dataset}&\textbf{Accuracy}\\
\hline
VGGFace &LFW \cite{LFW}& 98.95\%  \\

VGGFace &YTF \cite{YTF}&  97.3\% \\

VGGFace &FG-NET \cite{FGNET}&  84.9\% \\

VGGFace &CACD \cite{Chen:TransMult15}&  84.32\% \\
\hline
VGGFace &KANFace static& 75.81\% $\pm{0.6}$\\
\hline
VGGFace+GRU &KANFace video& 80.87\% $\pm{0.68}$\\
\hline

\end{tabular}

\caption{Baseline experimental results for face recognition. The second column contain the datasets that were used to fine-tune and evaluate the models.}
\label{tab:face_reco_im}
\end{table}

\subsubsection{Baseline experiments on age estimation}
\label{sub:age_esti}
To model the facial transformations caused by aging, we need a network trained on an extensive and diverse dataset. Our face age representation is trained on IMDb-Wiki \cite{Rothe:IJCV16}, the largest still image dataset with exact age labels. The image and video embeddings are extracted in a similar manner to those in the face recognition experiment. To obtain the age prediction, the embeddings are passed through a softmax layer with 101 neurons. Each output neuron is interpreted as the probability \textit{p(age)} of each age label, i.e., 0 to 100 years old, and the final prediction is calculated as the expected age: $o =\sum_{j=0}^{100} p(age=j)* j.$

\textbf{Evaluation protocol and baseline results: }Since our aim is to build a person-invariant age estimation system, open-set evaluation protocol is applied. That is, the subjects within the training set are excluded from the test set. In particular, the subject list is split into 5 folds, each consisting of 209 people. We perform 5-fold cross-validation on the corresponding images and videos, each time testing on a different set of identities. The method is evaluated based on the Mean Absolute Error (MAE), which is the standard metric for exact age estimation and is calculated as:  $MAE =\sum_{j=1}^N
|\bar{y}-y|/N.$
The mean MAE and standard deviation of our baseline experiments are reported in Table \ref{tab:age_esti_im}. The results are compared to standard publicly available age estimation benchmarks.

\begin{table}[hbt!]
\centering    
\begin{tabular}{|c|c|c|}

\hline
\textbf{Model}&\textbf{Dataset}&\textbf{MAE}\\
\hline
DEX-age &FG-NET& 3.09  \\

DEX-age &MORPH2 \cite{Morph}& 2.68  \\

DEX-age &CACD& 4.785  \\
\hline
DEX-age &KANFace static& 7.66 $\pm{0.08}$\\
\hline
DEX-age+GRU &KANFace video& 6.91 $\pm{0.44}$\\
\hline

\end{tabular}

\caption{Experimental results for age estimation. The second column contain the datasets that were used to fine-tune and evaluate the models.}
\label{tab:age_esti_im}
\end{table}

\subsubsection{Baseline experiments on gender recognition}
\label{sub:gender_reco}
To train the face representation for gender recognition, we utilize the gender annotations of the IMDb-Wiki dataset. Similarly to the above, the face and video descriptors are of 4,096 and 512 dimensions respectively. The descriptors are classified by a softmax classification layer with 2 output neurons.

\textbf{Evaluation protocol and baseline results: }We apply a person-invariant protocol and evaluate the model by conducting 5-fold cross-validation. The mean classification accuracy and standard deviation are reported in Table \ref{tab:gender_reco_im}. Since we study gender classification under aging transformations, the baseline model is also tested on MORPH 2, which has gender labels, using a random 80-20 split protocol.

\begin{table}[hbt!]
\centering    
\begin{tabular}{|c|c|c|}

\hline
\textbf{Model}&\textbf{Dataset}&\textbf{Accuracy}\\
\hline

DEX-gen &MORPH2&  96.7\% \\

\hline
DEX-gen &KANFace static& 93.04\% $\pm{0.72}$\\
\hline
DEX-gen+GRU &KANFace video& 96.12\% $\pm{1.22}$\\
\hline

\end{tabular}

\caption{Experimental results for gender recognition. The second column contain the datasets that were used to fine-tune and evaluate the models.}
\label{tab:gender_reco_im}
\end{table}

\subsubsection{Baseline experiments on kinship verification}\label{sub:base_kinship_ver}

For the task of kinship verification, we utilize a face representation trained on the VGGFace dataset and fine-tuned on the FIW dataset \cite{Robinson:FIW} using the triplet loss. The representation is further fine-tuned on CornellKin \cite{Fang:ICIP10}, KinFaceW-I \cite{Lu:TPAMI14} and KinFaceW-II \cite{Lu:TPAMI14}, and the results are presented on Table \ref{tab:kinship_ver_im}. For the experiments on videos, the VGGFace embedding is used as the face descriptor and the GRU is trained on the KANFace video dataset.

\textbf{Evaluation protocol and baseline results: }The proposed protocol focuses on the seven basic family relationships, i.e., Brother-Brother (B-B), Sister-Brother (S-B), Sister-Sister (S-S), Mother-Daughter (M-D), Mother-Son (M-S), Father-Daughter (F-D) and Father-Son (F-S). Open set evaluation is adopted, that is, 80\% of the subject pairs are used for training and 20\% for testing, with no subject overlap between the two sets. The negative pairs are generated randomly so that they are of the same gender as the corresponding positive pairs. The accuracy per relationship as well as the average accuracy of the baseline model is reported on Table \ref{tab:kinship_ver_im}.

 \begin{table*}[hbt!]
\centering    

\resizebox{\linewidth}{!}{
\begin{tabular}{|c|c|c|c|c|c|c|c|c|c|}

\hline
\textbf{Model}&\textbf{Dataset}&\textbf{B-B}& \textbf{S-S}&\textbf{S-B}& \textbf{M-S}& \textbf{M-D}&\textbf{F-S}& \textbf{F-D}&\textbf{Mean}\\
\hline
 VGG-FIW&CornellKin \cite{Fang:ICIP10}&-& -& -& -& -& -& -& 79.9\% \\
 VGG-FIW&KinFaceW-I \cite{Lu:TPAMI14}&-& -& -& 77.57\%& 79.96\%& 78.49\%& 79.16\% & 78.8\%\\
 VGG-FIW&KinFaceW-II \cite{Lu:TPAMI14}&-& -& -& 78.2\%& 79.4\%& 76.6\%& 73.8\% & 77\%\\
 \hline
 VGG-FIW&KANFace static&67.91\% & 66.69\% & 61.68\% & 59.57\% & 70.47\% & 68.08\% & 64.04\% &65.49\%  \\
 \hline
 VGGFace+GRU &KANFace video& 72.84\%&63.3\% &59.88\% & 80.1\% & 52.95\% & 63.38\%& 56.29\% &64.11\%\\
\hline

\end{tabular}
}

\centering

\caption{Experimental results for kinship verification. The second column shows the datasets that were used to fine-tune and evaluate the models.}
\label{tab:kinship_ver_im}
\end{table*}

\subsection{Discussion} \label{sub:baseline_discussion}

In Section \ref{div_analysis} we evaluated the diversity of the faces in the proposed KANFace dataset. In order to study how the diversity translates to model performance, the results of the baseline method are compared on different in-the-wild datasets in Sections \ref{sub:base_face_reco}-\ref{sub:base_kinship_ver}. Overall, the results on Tables \ref{tab:face_reco_im}-\ref{tab:kinship_ver_im} indicate that the large variation in age, illumination, occlusions and pose has resulted in a performance drop for all tasks.
A major factor for this performance gap is the age distribution of our dataset. Contrary to other datasets (e.g., \cite{Parkhi:BMVC15}), the proposed dataset contains a significant number of faces under 18 and over 60.
The large age variation proved to be challenging even when we trained the model on an aging dataset (\cite{Rothe:IJCV16}).

Furthermore, we notice a significant difference in the performance on the kinship verification task (\ref{sub:base_kinship_ver}). 
Besides the diverse nature of the proposed KANFace dataset, our benchmark presents a set of challenges that are not present in most previous works. 
Firstly, the proposed dataset is one of the largest kinship verification datasets, including 566,198 kin pairs (as opposed to 150 in CornellKin and 1000 in KinfaceW II). The large scale of the data comes with facial variations that are not present in smaller datasets.
Secondly, the problem we investigate introduces an age-invariant aspect to the task of kinship verification. For instance, we have test pairs where the parent is depicted at a much younger age than the child.
Lastly, the faces in our dataset are not cropped from the same image/video. This is vital for kinship analysis, as not doing so can induce  bias to the task.

\section{Bias analysis}\label{s:bias_analysis}

The experiments in the previous section indicate that the chosen baseline method performs poorly on the proposed KANFace dataset. Nevertheless, this fact alone is neither adequately informative nor transparent. In this work we take a different direction and attempt to diagnose the reason behind the poor performance of the model. We investigate the demographic bias of the model with regards to age and gender, using 5 age classes: (i) 0-18, (ii) 19-30, (iii) 31-45, (iv) 46-60 and (v) 61+. Additionally, we investigate the impact of illumination and skin-color tone (as measured by the ITA).

\subsection{Bias in face recognition}\label{s:bias_analysis_face_reco}

\begin{figure}[htp] 
    \centering
    \hspace{0.2\textwidth}

        \includegraphics[width=0.4\textwidth]{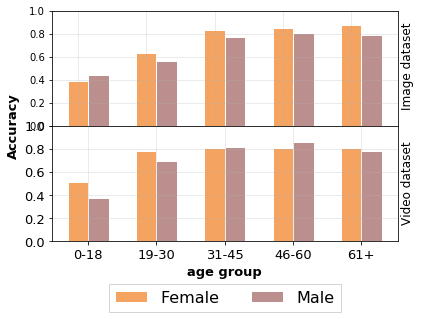}
    \caption{Analysis of performance of the chosen baseline model (see Section \ref{sub:base_face_reco}) for face recognition on images and videos.}
\label{fig:face_reco_bias}
\end{figure}

We investigate the performance of the chosen baseline face recognition model (see Section \ref{sub:base_face_reco}) and report the accuracy per age group and gender in Figure \ref{fig:face_reco_bias}. The results on the KANFace static images reveal that the performance is better on female faces (77.3\% accuracy for females and 75.1\% for males). The experiment on videos does not indicate such a bias, as the accuracy is similar for males and females (79.5\% accuracy for females and 80.4\% for males). Further to that, it is clear from Figure \ref{fig:face_reco_bias} that the model displays a bias towards the underrepresented age classes, e.g., 0-18 years old, where we notice a significant drop in performance.

In order to study how different models cope with the aforementioned biases, we experiment with different architectures. Besides the baseline model described in Section \ref{sub:base_face_reco}, we evaluate several deep models, including lightweight ones. In particular, we train LightCNN-9 \cite{Wu:TIFS18} and ResNet-50 \cite{Cao:VGGface2} on the KANFace image dataset and LSTM \cite{LSTM} on the KANFace video dataset. ResNet-50 is pretrained on VGGFace2 \cite{Cao:VGGface2} and fine-tuned on the proposed KANFace dataset, while LightCNN-9 is trained from scratch on KANFace. The results in Table  \ref{tab:bias_across_models_face_recognition} highlight the superior performance of ResNet-50. This is expected as this model is trained on \cite{Cao:VGGface2} which is larger and more diverse compared to \cite{Parkhi:BMVC15}. For the experiments on the video dataset, we train LSTM on KANFace in similar manner to GRU for comparison. We notice that GRU performs better in terms of average accuracy. However, the more complex LSTM model is less biased as indicated by its performance on faces under 18 and over 60.

\begin{table*}[hbt!]
\centering    

\resizebox{\linewidth}{!}{
\begin{tabular}{|c|c|c|c|c|c|c|c|c|c|c|c|}

\hline
\textbf{Model}&\textbf{F: 0-18}&\textbf{F: 19-30}& \textbf{F: 31-45}&\textbf{F: 46-60}& \textbf{F: 61+}&\textbf{M: 0-18}&\textbf{M: 19-30}& \textbf{M: 31-45}&\textbf{M: 46-60}& \textbf{M: 61+}& \textbf{Acc}\\
\hline
 LightCNN-9\cite{Wu:TIFS18}&0.215&0.394& 0.522& 0.58& 0.67& 0.19& 0.263& 0.505 & 0.593& 0.594&0.5055\\
 VGG-16\cite{Parkhi:BMVC15}&0.395&0.632& \textbf{0.837}& 0.852& 0.883& 0.448& 0.568& 0.775& 0.813 &0.794&0.7618\\
 ResNet-50\cite{Cao:VGGface2}&\textbf{0.556}&\textbf{0.688}& 0.835& \textbf{0.856}& \textbf{0.935}& \textbf{0.469}& \textbf{0.662}& \textbf{0.833} &\textbf{ 0.842}&\textbf{0.825}&\textbf{0.8008}\\
 \hline
VGG-16-GRU&0.52&\textbf{0.781}& \textbf{0.815}& \textbf{0.812}& 0.811& 0.376& \textbf{0.695}& \textbf{0.82}& \textbf{0.866} &0.789&\textbf{0.8006}\\
VGG-16-LSTM&\textbf{0.568}&0.718&0.797&0.784&\textbf{0.832}&\textbf{0.471}&0.671&\textbf{0.82}&0.857&\textbf{0.791}&0.791\\

\hline

\end{tabular}
}

\centering

\caption{Experimental results for face recognition on images and videos. F: Female, M: Male. The models are trained/fine-tuned and evaluated on KANFace.}
\label{tab:bias_across_models_face_recognition}
\end{table*}

\subsection{Bias in age estimation}\label{s:bias_analysis_age_esti}

\begin{figure}[htp] 
    \centering
    \hspace{0.2\textwidth}

        \includegraphics[width=0.4\textwidth]{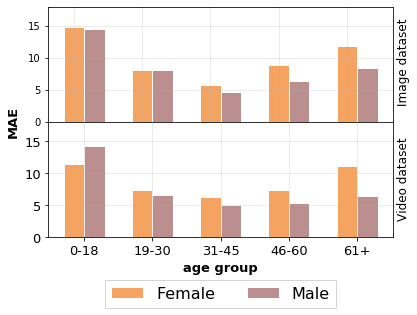}
    \caption{Analysis of performance of the chosen baseline model (see Section \ref{sub:age_esti}) for age estimation on images and videos.}
\label{fig:age_esti_bias}
\end{figure}

A similar analysis is conducted for the chosen baseline age estimation model described in Section \ref{sub:age_esti}. The results in Figure \ref{fig:age_esti_bias} reveal that the model is biased towards the tails of the age distribution. In particular, MAE increases significantly for faces under 18 and over 60. This is due to the age distribution of the training set (i.e., \cite{Rothe:IJCV16}) as well as the proposed dataset (Figure \ref{fig:ages}).

Since the proposed age estimation protocol is person invariant, we are able to evaluate bias not only across models but also across datasets. We perform cross-dataset experiments on FG-NET, MORPH 2 and KANFace. Along with the baseline model, we study the lightweight SSR-NET \cite{Yang:IJCAI18} architecture. With orders of magnitude less parameters than VGG-16, this lightweight model is compact and cannot perform on par with our baseline. Nevertheless, the comparison offers insights regarding the effect of model complexity on bias. We compare LSTM and GRU on the video dataset similarly to the previous Section.

Overall, compared to training on FG-NET or MORPH 2, the models that are trained on KANFace perform better across datasets (Figure \ref{fig:age_analysis}). Moreover, using a more complex model that is pretrained on a rich datasets such as \cite{Rothe:IJCV16} yields better and more fair results, especially when tested on a different dataset.
We analyze the performance of all trained models as shown in Table \ref{tab:bias_across_models_age_estimation}. The results highlight the effect of the diversity of the training set, concretely: 1) Since more than 68\% of FG-NET consists of faces between 0 and 18 years old, the models that are trained on FG-NET perform significantly better on that population. 2) The models that are trained on MORPH 2 and the KANFace dataset perform better on faces between 19 and 45 years old. Indicatively, more than 75\% of the faces in MORPH 2 belong to this age group. 3) Only the models that are trained on KANFace are able to perform adequately on faces over 45 years old, as they represent more than 40\% of the  faces in the dataset. The experimental results on the video dataset indicate that the GRU and LSTM perform similarly, with the latter performing marginally better on average.

\begin{table*}[hbt!]
\centering    

\resizebox{\linewidth}{!}{
\begin{tabular}{|c|c|c|c|c|c|c|c|c|c|c|c|c|}

\hline
\textbf{Model- training set}&\textbf{F: 0-18}&\textbf{F: 19-30}& \textbf{F: 31-45}&\textbf{F: 46-60}& \textbf{F: 61+}&\textbf{M: 0-18}&\textbf{M: 19-30}& \textbf{M: 31-45}&\textbf{M: 46-60}& \textbf{M: 61+}& \textbf{MAE}\\
\hline
SSRNET-FGNET&6.2&11.52&21.69&33.06&47.68&5.04&10.37&19.98&27.99&38.76 & 24.08\\
SSRNET-MORPH&25.46&16.5&10.21&10.41& 25.34& 30.0&  19.28& 9.95& 7.63 &21.52& 14.27\\
SSRNET-KANFace&22.12&11.15&7.06&11.21&20.29&24.83&12.88&8.08&8.34&12.63& 11.04\\
VGG16-FGNET&\textbf{4.22}&7.96& 14.72& 22.48& 23.3& \textbf{6.26}& 8.46&13.92 & 18.38 &19.1 & 15.17\\
VGG16-MORPH&15.59&7.49&\textbf{5.73}&10.51&19.92&19.34&\textbf{7.72}&\textbf{5.19}&8.31&16.62& 9.47\\
VGG16-KANFace&13.09&\textbf{7.13}& 5.96& \textbf{9.13}& \textbf{11.39}& 14.15& 7.84& 5.38& \textbf{7.0} &\textbf{8.87} & \textbf{7.65}\\
\hline
VGG16-GRU&\textbf{11.43}&\textbf{7.42}&6.36&\textbf{7.38}&11.14&14.22&\textbf{6.61}&5.12&\textbf{5.41}&\textbf{6.46}& 6.76\\
VGG16-LSTM&11.64&7.56&\textbf{5.79}&8.98&\textbf{10.8}&\textbf{13.58}&6.78&\textbf{4.58}&5.72&6.55 & \textbf{6.75}\\
\hline

\end{tabular}
}

\centering

\caption{Bias analysis of the age estimation models. F: Female, M: Male. The models were evaluated on KANFace.}

\label{tab:bias_across_models_age_estimation}
\end{table*}

\subsection{Bias in gender recognition}\label{s:bias_analysis_gender_recognition}
\begin{figure}[H] 
    \centering
    \hspace{0.2\textwidth}

        \includegraphics[width=0.4\textwidth]{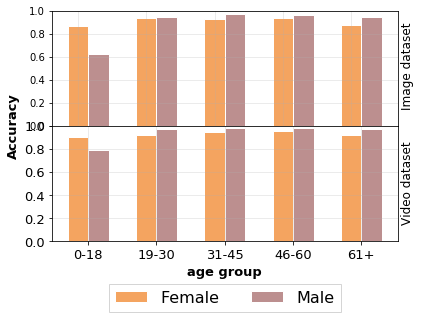}
    \caption{Analysis of performance of baseline models for gender recognition on images and videos.}
\label{fig:gender_reco_bias}
\end{figure}

The results in Table \ref{tab:gender_reco_im} indicate that the gender recognition task does not pose a significant challenge to our baseline models. Nevertheless, by analyzing the performance of the models (Figure \ref{fig:gender_reco_bias}) we are able to uncover biased behavior towards male faces under 18 years old. The results agree with studies that support that gender recognition of male faces is facilitated by masculinity \cite{role_of_attractiveness_masculinity}, which is largely correlated to attributes that appear in later stages of facial development.

\begin{figure*}[!b] 
    \centering

    \subfloat{%
         \includegraphics[width=0.9\textwidth]{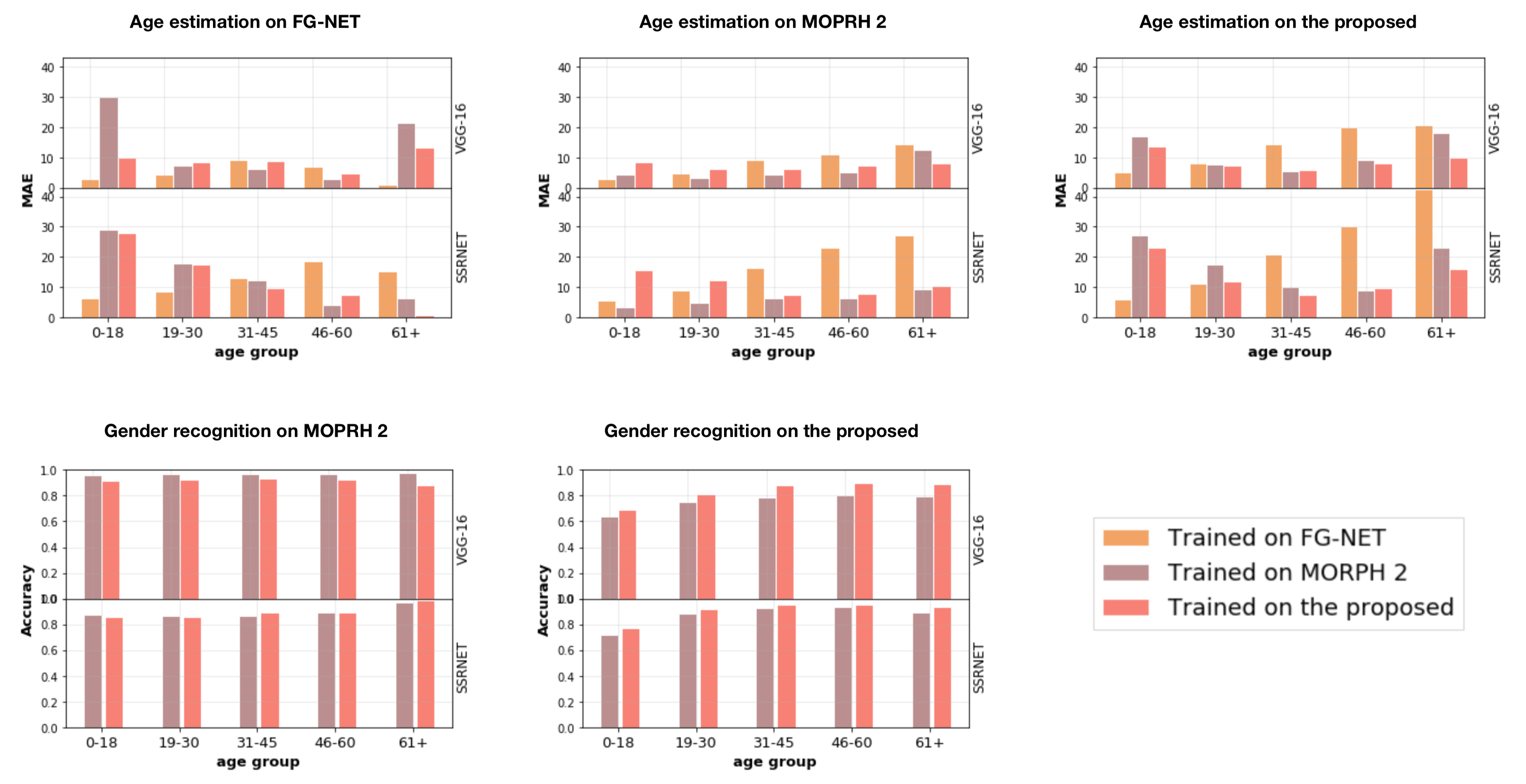}%
         \label{fig:age_cros_morph}%
         }%
        
    \caption{Bias analysis of the cross-dataset age estimation and gender recognition experiments. The different colors denote the different training sets.}
    \label{fig:age_analysis}
\end{figure*}

Similarly to Section \ref{s:bias_analysis_age_esti}, cross-dataset experiments are conducted on KANFace and MORPH 2 and presented in Table \ref{tab:bias_analysis_gender} and Figure \ref{fig:age_analysis}. The models that are trained on MORPH 2 demonstrate a significant performance drop when tested on a different dataset. Indicatively, for the SSR-NET the performance drops from 87.4\% to 77.7\% and for the  baseline VGG-16 from 96.8\% to 90.4\%. When trained on the proposed KANFace dataset, the performance remains similar across datasets (86.32\% to 87.67\% for SSR-NET and 93.5\% to 93\% for VGG-16). Lastly, all models displayed the aforementioned age bias towards male faces. However, the much smaller SSR-NET is the least biased, when trained on KANFace.

\begin{table*}[hbt!]
\centering    

\resizebox{\linewidth}{!}{
\begin{tabular}{|c|c|c|c|c|c|c|c|c|c|c|c|c|c|}

\hline
\textbf{Model- training set}&\textbf{F: 0-18}&\textbf{F: 19-30}& \textbf{F: 31-45}&\textbf{F: 46-60}& \textbf{F: 61+}&\textbf{M: 0-18}&\textbf{M: 19-30}& \textbf{M: 31-45}&\textbf{M: 46-60}& \textbf{M: 61+} & \textbf{Acc.}\\
\hline

SSRNET-MORPH&0.705&0.731&0.764&0.766&0.734&0.54&0.776&0.816&0.827&0.84&0.777\\
SSRNET-KANFace&0.684&0.765&0.83&0.866&0.861&\textbf{0.693}&0.877&0.935&0.924&0.905&0.8632\\
VGG16-MORPH&0.816&0.878&0.903&0.917&0.918&0.56&\textbf{0.897}&\textbf{0.952}&0.949&0.869&0.904\\
VGG16-KANFace&\textbf{0.923}&\textbf{0.961}&\textbf{0.981}&\textbf{0.961}&\textbf{0.964}&0.527&0.843&0.927&\textbf{0.951}&\textbf{0.923}&\textbf{0.935}\\
\hline
VGG16-GRU&0.905&0.924&0.954&0.955&0.921&\textbf{0.797}&\textbf{0.975}&\textbf{0.986}&\textbf{0.981}&\textbf{0.976}&\textbf{0.959}\\
VGG16-LSTM&\textbf{0.942}&\textbf{0.944}&\textbf{0.969}&\textbf{0.974}&\textbf{0.939}&0.632&0.951&0.975&0.961&0.961&\textbf{0.959}\\
\hline

\end{tabular}
}

\centering

\caption{Bias analysis of the gender recognition models. F: Female, M: Male. The models were evaluated on KANFace.}
\label{tab:bias_analysis_gender}
\end{table*}

\subsection{Bias in kinship verification}\label{s:bias_analysis_kin_ver}

\begin{figure*}[htp] 
    \centering

        \includegraphics[width=0.9\textwidth]{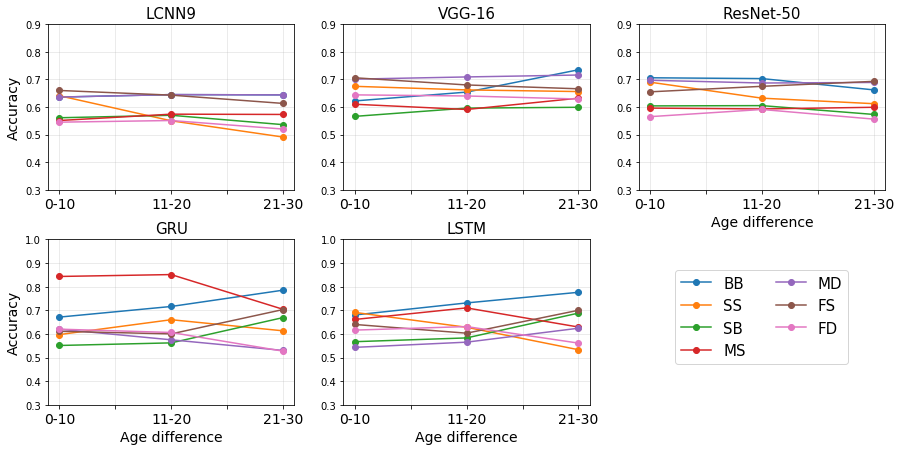}%

    \caption{Bias analysis of the kinship verification models with regards to age difference. The models were trained/fine-tuned on and evaluated on KANFace. BB: Brother-Brother, SS: Sister-Sister, SB: Siblings, MS: Mother-Son, MD: Mother-Daughter, FS: Father-Son, FD: Father-Daughter}
    \label{fig:kinship_analysis}
\end{figure*}

As discussed in Section \ref{sub:baseline_discussion}, the proposed kinship verification benchmark presents the novel challenge in terms of large variation in the age difference of the input faces. Motivated by this, we study the performance of the baseline models described in Section \ref{sub:base_kinship_ver} for three ranges of age difference, namely 0-10 years, 11-20 years and 21-30 years. Besides the baseline models, we conduct experiments with ResNet-50 and LightCNN-9 on images, as well as with LSTM on the video dataset. The results are presented in Figure \ref{fig:kinship_analysis}.
Following common practices (e.g. \cite{FIW}), the image-based kinship verification models were pretrained on face recognition. Hence, the resulting models should encode the age-invariant features that are vital for face recognition. Indeed, the results on Figure \ref{fig:kinship_analysis} do not indicate significant change in performance with varying age difference. The performance of the temporal models varies more with age difference, with the LSTM performing marginally better on average.

\subsection{Bias with regards to individual typology angle}\label{s:bias_analysis_ita}

\begin{figure*}[!b] 
    \centering

    \subfloat[Face recognition]{%
        \includegraphics[width=0.3\textwidth]{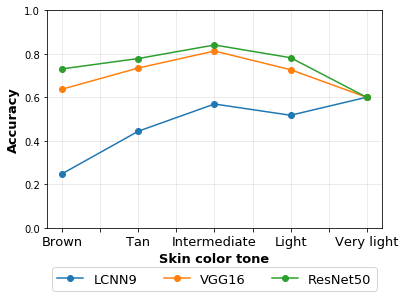}%

        }%
    \hfill%
    \subfloat[Age estimation]{%
        \includegraphics[width=0.35\textwidth]{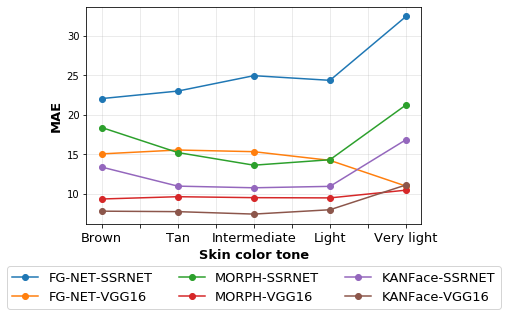}%

        }%
        \hfill%
    \subfloat[Gender recognition]{%
        \includegraphics[width=0.27\textwidth]{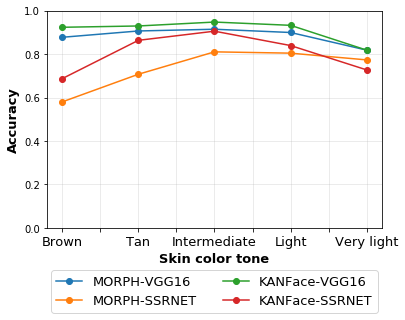}%

        }
        \caption{Bias analysis of the results with regards to ITA. The model were evaluated on KANFace. Face recognition models (a) were trained/fine-tuned on KANFace. Age estimation (b) and gender recognition (c) models were trained/fine-tuned the specified dataset.}
    \label{fig:ita_analysis}
\end{figure*}

To study the effect of perceived skin-color tone on the performance of the trained models, we perform an analysis with regards to perceived skin color. We focus on five individual typology angle classes, namely: -30-10 (brown), 11-28 (tan), 29-41 (intermediate), 24-55 (light) and over 55 (very light). The results for the tasks of face recognition, age estimation and gender recognition are presented in Figure \ref{fig:ita_analysis}. In general, the models perform better on intermediate faces, with the performance dropping as the faces get lighter or darker. Pretraining the model on a large and diverse dataset also seems to help with generalization across perceived skin-color tone, as the models that are trained on \cite{Simonyan:VGG16}, \cite{Cao:VGGface2} and \cite{Rothe:IJCV16} (i.e., VGG16 and ResNet50) perform better across ITA classes and are in general more less biased.

\section{Mitigation of representation bias}\label{s:debias}

In the previous section, we presented a series of experimental results that reveal the existence of bias in a series of face anlysis tasks. We recognize the demographic imbalance of the training set as the source of bias in the learned representation, which in turn results in biased classification. In order to debias the deep representation, we propose to decompose a network embedding $z\in \mathbb{R}^{d_1}$ as follows:
\begin{equation}
z  = z_p + \sum_{i=0}^N z_i, 
\end{equation}
where  $z_p$ is the representation that is discriminative for the primary classification task (e.g., face recognition) and $z_i$ are the representations for the sensitive attributes (e.g., age and gender). It is clear that we want the representations for the sensitive attributes to be independent of the primary representation, i.e., $z_p \independent z_i$ for every sensitive attribute  $i=1,\dots,N$. In order to obtain the debiased representation while imposing the independence of the components, we further assume the following decomposition:

\begin{align}
&z  = A\; B\; z + \sum_{i=0}^N D_i\; T_i \; B \; z, \\
s.&t.\; A^{T} \;D_i = 0, \;n =0,1,\dots,N \nonumber
\end{align}
where $A\in \mathbb{R}^{d_1\times d_2}, B\in \mathbb{R}^{d_2\times d_1}, T_i\in \mathbb{R}^{d_3\times d_2}, D_i\in \mathbb{R}^{d_1\times d_2}$ and solve the following optimization problem:
\begin{equation}
\min\limits_{V} \max\limits_{U} \mathcal{L}_{cls} + \lambda_{dec} \mathcal{L}_{decom} + \lambda_{or} \mathcal{L}_{or} -  \sum_{i=0}^N \lambda_{i} \mathcal{L}_{entr}^{i} .
\label{eq:minmax}
\end{equation}
The parameter sets $V$ and $U$ are defined as $V = \{A,B,D_i, W_p\}$ and $U = \{T_i, W_i\}$. To solve \ref{eq:minmax}, we formulate the optimization problem using adversarial learning and solve the following subproblems:

\begin{equation}
\min\limits_{V}
\mathcal{L}_{cls}^{p} + \lambda_{dec} \mathcal{L}_{decom} + \mathcal{L}_{or} - \sum_{i=0}^N\lambda_{i}\mathcal{L}_{entr}^{i} 
\end{equation}

and

\begin{equation}
\min\limits_{U}
\sum_{i=0}^N \mathcal{L}_{cls}^{i}
\end{equation}
where $\mathcal{L}_{cls}^{p}$ and $\mathcal{L}_{cls}^{i}$ are the softmax classification losses with parameters $W_p$ and $W_i$ and $\mathcal{L}_{entr}^{i}$ is the entropy of the classifier for the $i-th$ sensitive attribute. Lastly, the decomposition and orthogonality losses are defined as: 
\begin{align}
    \mathcal{L}_{decom} &=  \frac{1}{2}\lVert z - A\; B\;
    z - \sum_{i=0}^N D_i\; T_i \; B \; z \lVert_F^2\\
    \mathcal{L}_{or} &=  \sum_{i=0}^N \lambda_{or}^i\;\frac{1}{2}\lVert  A^T\; D_i \lVert_F^2
\end{align}.

We apply the proposed method on the embeddings of our baseline VGG-16 and GRU classification models (i.e., face recognition, age estimation and gender recognition) on images and videos. Our goal is to investigate when unwanted demographic information can be disentangled from the representation without losing discriminative power. The sensitive attributes for each task are chosen based on the bias analysis of the previous section. Similar to the baseline experiments, the models were fine-tuned and evaluated on the proposed KANFace dataset.

The proposed debiasing method leverages adversarial learning to impose the independence between the representations of the primary task and the sensitive attributes. Similar approaches have been proposed in the domain adaptation literature (e.g., \cite{ganin2016domain, tzeng2015simultaneous, tzeng2017adversarial}) in order to minimize the domain shift. Contrary to such methods, our framework is used on pretrained embeddings and extracts representations for both the primary task and the sensitive attributes.

\subsection{Debiased representation for face recognition}

The face recognition baselines were biased with regards to age and gender (Figure \ref{fig:face_reco_bias}). Therefore, we apply the proposed method to debias the image and video embeddings, setting age and gender as sensitives attributes. The results in Figure \ref{fig:debias} show that the obtained representations perform better for the underrepresented classes of under 18 and over 60 years old. Furthermore, the new classifier is less biased across genders. Compared to the video features, the image features benefited the most from the proposed debiasing method. This is due to the dimensionality of the image features (image features are 4,096-dimensional, while video features are 512-dimensional) and indicate that high dimensional deep features run the risk of encoding bias-inducing information.

\begin{figure}[] 
    \centering

        \includegraphics[width=0.45\textwidth]{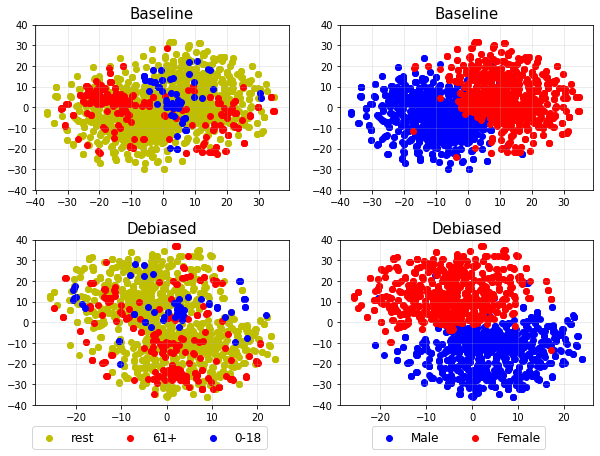}
    \caption{t-SNE of the baseline and debiased face recognition embeddings. The figures on the left column represent the underrepresented age classes. The faces under 18 (in blue) are significantly less clustered. The figures on the right represent the gender classes.}
\label{fig:face_reco_tsne}
\end{figure}

Furthermore, we investigate whether age and gender information can be disentangled from face recognition features using the proposed method. To this end, we use t-SNE \cite{tsne} to analyze the baseline and debiased representations. The results on Figure \ref{fig:face_reco_tsne} highlight the ability of the proposed method to disentangle age related information, especially for faces under 18 years old. However, the same cannot be said for gender, as both the baseline and debiased representations clearly encode gender information. This result is consistent with psychology studies that suggest that the gender of a face is highly correlated with the perceived identity \cite{gender_is_a_dimension}. Hence, the proposed decomposition could not disentangle gender information from the representation without decreasing face recognition accuracy.

\subsection{Debiased representation for age estimation}

The bias analysis of the image and video age estimation models (Figure \ref{fig:age_esti_bias}) indicate the existence of gender bias, as both the image and video baseline models perform better on male faces. Furthermore, the difference in MAE between male and female faces increases with the age of the face. Psychological studies suggest that this bias is due to the correlation of perceived age with perceived attractiveness \cite{let_guess_age}. In these studies, women exhibit larger variability in their efforts to look younger, hence inducing bias in age estimation. Since the suggested dataset consists of celebrity faces, such bias is dominant, with the existence of make-up or plastic surgeries inducing noise to age prediction. Therefore, as indicated by the results on Figure \ref{fig:debias_age}, our method is not able to mitigate gender bias adequately.

\subsection{Debiased representation for gender recognition}

\begin{figure*}[!] 
    \centering

    \subfloat[Face recognition]{%
        \includegraphics[width=0.3\textwidth]{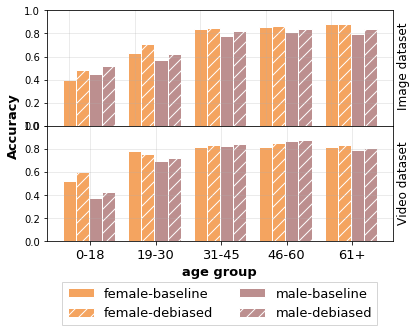}%
        \label{fig:debias_face}%
        }%
    \hfill%
    \subfloat[Age estimation]{%
        \includegraphics[width=0.3\textwidth]{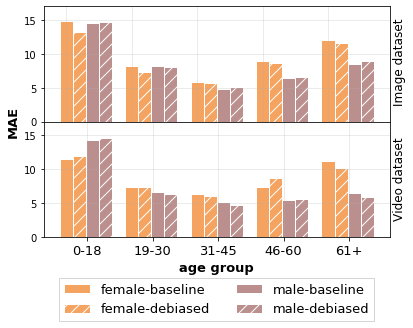}%
        \label{fig:debias_age}%
        }%
        \hfill%
    \subfloat[Gender recognition]{%
        \includegraphics[width=0.3\textwidth]{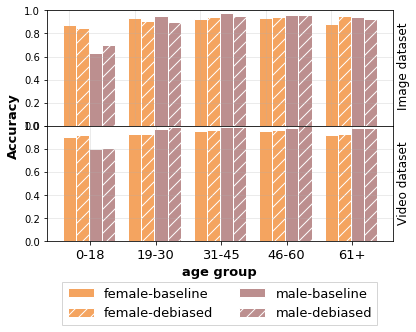}%
        \label{fig:debias_gender}%
        }
    \caption{Performance comparison between the baseline representations for face recognition (see Section \ref{sub:base_face_reco}), age estimation (see Section \ref{sub:age_esti}) and gender recognition (see Section \ref{sub:gender_reco}) and the debiased representations produced by the proposed method for each model. All models were trained and evaluated on KANFace. The proposed method is able to remove the bias for face recognition (a) and gender recognition (c), but it is not able to remove the gender bias in age estimation (b).}
    \label{fig:debias}
\end{figure*}

The analysis in Section \ref{s:bias_analysis_gender_recognition} revealed that the gender recognition baseline models are biased towards male faces under 18 years old. In order to mitigate age bias, we apply the proposed method on the image and video embeddings. The results on Figure \ref{fig:debias_gender} show that the debiased representations perform better on male faces under 18 years old, especially on images. The lower dimensional video representation did not display the aforementioned bias to the same extent, and therefore did not benefit as much from the debiasing.

\section{Conclusion}

Deep neural networks have been successfully used to advance the state-of-the-art in image and video-based face analysis by capitalizing on the existence of a large corpus of available facial datasets. However, the lack of diversity in such datasets resulted in biased models. In this work we introduce KANFace dataset, the largest manually annotated image and video dataset for face analysis in-the-wild. Being thoroughly annotated with regards to age, gender and kinship, the proposed dataset allows for the study of novel tasks such as video-based age estimation and kinship verification.

Making use of the rich annotations, we investigate the bias in state-of-the-art deep learning models on a variety of image and video-based face analysis tasks, namely face recognition, age estimation, gender recognition and kinship verification. To the best of our knowledge, this is the first work to put the bias of all these tasks under scrutiny. 

Our extensive experimental evaluation revealed biased behaviour towards certain demographics and provided insights of why this is happening. In particular, both face recognition and age estimation models were biased towards faces under 18 and over 60 years old. The age estimation models were also biased towards female faces, especially older ones, due to the presence of nuisance factors that appear on celebrity faces, e.g., make-up and plastic surgery. On the other hand, gender recognition models displayed age bias towards young male faces, possibly due to the lack of features that are developed on adult male faces. Lastly, the experiments on kinship verification indicate that, despite the inherent challenges of age-invariant kinship modeling, pretraining the models for face recognition yields age-invariant kinship representations.

We introduce a method to debias the network embeddings. We apply the proposed method to investigate in which cases demographic information can be disentangled from the deep representation. The results indicated that the age bias that exists in face recognition and gender recognition can be mitigated. On the other hand, gender bias cannot be mitigated as it a discriminative attribute for face recognition. Similarly, we observe that the proposed method cannot eradicate gender bias from the age estimation embeddings.
The proposed bias analysis protocol can be applied to diagnose the performance of any face analysis model. Thus, both KANFace dataset and the bias analysis protocol presented in this work constitute tools towards unbiased face analysis models from still images and videos in-the-wild.

\bibliographystyle{elsarticle-num} 
\bibliography{egbib}

\begin{thebibliography}{10}
\expandafter\ifx\csname url\endcsname\relax
  \def\url#1{\texttt{#1}}\fi
\expandafter\ifx\csname urlprefix\endcsname\relax\def\urlprefix{URL }\fi
\expandafter\ifx\csname href\endcsname\relax
  \def\href#1#2{#2} \def\path#1{#1}\fi

\bibitem{masi2018deep}
I.~Masi, Y.~Wu, T.~Hassner, P.~Natarajan, Deep face recognition: A survey, in:
  2018 31st SIBGRAPI conference on graphics, patterns and images (SIBGRAPI),
  IEEE, pp. 471--478.

\bibitem{Ranjan:Hyperface}
R.~Ranjan, V.~M. Patel, R.~Chellappa, Hyperface: A deep multi-task learning
  framework for face detection, landmark localization, pose estimation, and
  gender recognition, IEEE Transactions on Pattern Analysis and Machine
  Intelligence (2017).

\bibitem{vougioukas2019realistic}
K.~Vougioukas, S.~Petridis, M.~Pantic, Realistic speech-driven facial animation
  with gans, International Journal of Computer Vision (2019) 1--16.

\bibitem{siarohin2019first}
A.~Siarohin, S.~Lathuili{\`e}re, S.~Tulyakov, E.~Ricci, N.~Sebe, First order
  motion model for image animation, in: Advances in Neural Information
  Processing Systems, 2019, pp. 7135--7145.

\bibitem{liu2020deep}
L.~Liu, W.~Ouyang, X.~Wang, P.~Fieguth, J.~Chen, X.~Liu, M.~Pietik{\"a}inen,
  Deep learning for generic object detection: A survey, International journal
  of computer vision 128~(2) (2020) 261--318.

\bibitem{minaee2020image}
S.~Minaee, Y.~Boykov, F.~Porikli, A.~Plaza, N.~Kehtarnavaz, D.~Terzopoulos,
  Image segmentation using deep learning: A survey, arXiv preprint
  arXiv:2001.05566 (2020).

\bibitem{pie}
T.~Sim, S.~Baker, M.~Bsat, The cmu pose, illumination, and expression (pie)
  database, in: IEEE International Conference on Automatic Face and Gesture
  Recognition (FG), IEEE, 2002, pp. 53--58.

\bibitem{LFW}
G.~B. Huang, M.~Ramesh, T.~Berg, E.~Learned-Miller, Labeled faces in the wild:
  A database for studying face recognition in unconstrained environments, Tech.
  rep.

\bibitem{Megaface}
I.~Kemelmacher-Shlizerman, S.~M. Seitz, D.~Miller, E.~Brossard, The megaface
  benchmark: 1 million faces for recognition at scale, in: Proceedings of the
  IEEE Conference on Computer Vision and Pattern Recognition, 2016, pp.
  4873--4882.

\bibitem{Turk:Eigenfaces}
M.~A. Turk, A.~P. Pentland, Face recognition using eigenfaces, in: IEEE
  Conference on Computer Vision and Pattern Recognition (CVPR), IEEE, 1991, pp.
  586--591.

\bibitem{torralba:dataset}
A.~Torralba, A.~A. Efros, et~al., Unbiased look at dataset bias., Citeseer.

\bibitem{gender_shades}
J.~Buolamwini, T.~Gebru, Gender shades: Intersectional accuracy disparities in
  commercial gender classification, in: Conference on Fairness, Accountability
  and Transparency, 2018, pp. 77--91.

\bibitem{Celeba}
Z.~Liu, P.~Luo, X.~Wang, X.~Tang, Deep learning face attributes in the wild,
  in: Proceedings of International Conference on Computer Vision (ICCV), 2015.

\bibitem{Cao:VGGface2}
Q.~Cao, L.~Shen, W.~Xie, O.~M. Parkhi, A.~Zisserman, Vggface2: A dataset for
  recognising faces across pose and age, in: International Conference on
  Automatic Face and Gesture Recognition, 2018.

\bibitem{Rothe:IJCV16}
R.~Rothe, R.~Timofte, L.~V. Gool, Deep expectation of real and apparent age
  from a single image without facial landmarks, International Journal of
  Computer Vision (July 2016).

\bibitem{ownage_bias}
A.~Schaich, S.~Obermeyer, T.~Kolling, M.~Knopf, An own-age bias in recognizing
  faces with horizontal information, Frontiers in aging neuroscience 8 (2016)
  264.

\bibitem{owngender_bias}
J.~Lov{\'e}n, A.~Herlitz, J.~Rehnman, Women’s own-gender bias in face
  recognition memory, Experimental psychology (2011).

\bibitem{ownrace_bias}
R.~K. Bothwell, J.~C. Brigham, R.~S. Malpass, Cross-racial identification,
  Personality and Social Psychology Bulletin 15~(1) (1989) 19--25.

\bibitem{Zou:Nature}
J.~Zou, L.~Schiebinger, Ai can be sexist and racist—it’s time to make it
  fair (2018).

\bibitem{Lopez:TPAMI16}
M.~B. Lopez, E.~Boutellaa, A.~Hadid, Comments on the kinship face in the wild
  data sets, IEEE Transactions on Pattern Analysis and Machine Intelligence
  38~(11) (2016) 2342--2344.
\newblock \href {https://doi.org/10.1109/TPAMI.2016.2522416}
  {\path{doi:10.1109/TPAMI.2016.2522416}}.

\bibitem{Georgopoulos:IMAVIS18}
M.~Georgopoulos, Y.~Panagakis, M.~Pantic, Modeling of facial aging and kinship:
  A survey, Image and Vision Computing (2018).

\bibitem{tommasi2017deeper}
T.~Tommasi, N.~Patricia, B.~Caputo, T.~Tuytelaars, A deeper look at dataset
  bias, in: Domain adaptation in computer vision applications, Springer, 2017,
  pp. 37--55.

\bibitem{DiF}
M.~Merler, N.~Ratha, R.~S. Feris, J.~R. Smith, Diversity in faces, arXiv
  preprint arXiv:1901.10436 (2019).

\bibitem{FGNET}
A.~Lanitis, F{G-NET} {A}ging {D}atabase (2002).

\bibitem{Lanitis:TPAMI02prog}
A.~Lanitis, C.~J. Taylor, T.~F. Cootes, Toward automatic simulation of aging
  effects on face images, IEEE Transactions on Pattern Analysis and Machine
  Intelligence 24~(4) (2002) 442--455.

\bibitem{Morph}
K.~Ricanek, T.~Tesafaye, Morph: A longitudinal image database of normal adult
  age-progression, in: IEEE International Conference on Automatic Face and
  Gesture Recognition (FG), IEEE, 2006, pp. 341--345.

\bibitem{Fang:ICIP10}
R.~Fang, K.~D. Tang, N.~Snavely, T.~Chen, Towards computational models of
  kinship verification, in: IEEE International Conference on Image Processing
  (ICIP), IEEE, 2010, pp. 1577--1580.

\bibitem{YTF}
L.~Wolf, T.~Hassner, I.~Maoz, Face recognition in unconstrained videos with
  matched background similarity, in: Computer Vision and Pattern Recognition
  (CVPR), 2011 IEEE Conference on, IEEE, 2011, pp. 529--534.

\bibitem{Xia:IJCAI11}
S.~Xia, M.~Shao, Y.~Fu, Kinship verification through transfer learning, in:
  IJCAI Proceedings-international joint conference on artificial intelligence,
  Vol.~22, 2011, p. 2539.

\bibitem{Shao:CVPRw11}
M.~Shao, S.~Xia, Y.~Fu, Genealogical face recognition based on ub kinface
  database, in: IEEE Conference on Computer Vision and Pattern Recognition
  Workshop (CVPR- W), IEEE, 2011, pp. 60--65.

\bibitem{Dibekliouglu:ECCV12}
H.~Dibeklio{\u{g}}lu, A.~Salah, T.~Gevers, Are you really smiling at me?
  spontaneous versus posed enjoyment smiles, European Conference on Computer
  Vision (ECCV) (2012) 525--538.

\bibitem{Eidinger:TIFS14}
E.~Eidinger, R.~Enbar, T.~Hassner, Age and gender estimation of unfiltered
  faces, IEEE Transactions on Information Forensics and Security 9~(12) (2014)
  2170--2179.
\newblock \href {https://doi.org/10.1109/TIFS.2014.2359646}
  {\path{doi:10.1109/TIFS.2014.2359646}}.

\bibitem{Chen:TransMult15}
B.~C. Chen, C.~S. Chen, W.~H. Hsu, Face recognition and retrieval using
  cross-age reference coding with cross-age celebrity dataset, IEEE
  Transactions on Multimedia 17~(6) (2015) 804--815.
\newblock \href {https://doi.org/10.1109/TMM.2015.2420374}
  {\path{doi:10.1109/TMM.2015.2420374}}.

\bibitem{casia_webface}
D.~Yi, Z.~Lei, S.~Liao, S.~Z. Li, Learning face representation from scratch,
  arXiv preprint arXiv:1411.7923 (2014).

\bibitem{Lu:TPAMI14}
J.~Lu, X.~Zhou, Y.-P. Tan, Y.~Shang, J.~Zhou, Neighborhood repulsed metric
  learning for kinship verification, IEEE transactions on pattern analysis and
  machine intelligence 36~(2) (2014) 331--345.

\bibitem{Parkhi:BMVC15}
O.~M. Parkhi, A.~Vedaldi, A.~Zisserman, Deep face recognition, in: British
  Machine Vision Conference, 2015.

\bibitem{UMDfaces}
A.~Bansal, A.~Nanduri, C.~D. Castillo, R.~Ranjan, R.~Chellappa, Umdfaces: An
  annotated face dataset for training deep networks, in: Biometrics (IJCB),
  2017 IEEE International Joint Conference on, IEEE, 2017, pp. 464--473.

\bibitem{Bansal:ICCVW17}
A.~Bansal, C.~Castillo, R.~Ranjan, R.~Chellappa, The do’s and don’ts for
  cnn-based face verification, in: 2017 IEEE International Conference on
  Computer Vision Workshop (ICCVW), IEEE, 2017, pp. 2545--2554.

\bibitem{msceleb1m}
Y.~Guo, L.~Zhang, Y.~Hu, X.~He, J.~Gao, Ms-celeb-1m: A dataset and benchmark
  for large-scale face recognition, in: European Conference on Computer Vision,
  Springer, 2016, pp. 87--102.

\bibitem{FIW}
J.~P. Robinson, M.~Shao, Y.~Wu, Y.~Fu,
  \href{http://doi.acm.org/10.1145/2964284.2967219}{Families in the wild (fiw):
  Large-scale kinship image database and benchmarks}, in: Proceedings of the
  2016 ACM on Multimedia Conference, MM '16, ACM, New York, NY, USA, 2016, pp.
  242--246.
\newblock \href {https://doi.org/10.1145/2964284.2967219}
  {\path{doi:10.1145/2964284.2967219}}.
\newline\urlprefix\url{http://doi.acm.org/10.1145/2964284.2967219}

\bibitem{robinson2018visual}
J.~P. Robinson, M.~Shao, Y.~Wu, H.~Liu, T.~Gillis, Y.~Fu, Visual kinship
  recognition of families in the wild, IEEE Transactions on pattern analysis
  and machine intelligence 40~(11) (2018) 2624--2637.

\bibitem{IJB-B}
C.~Whitelam, E.~Taborsky, A.~Blanton, B.~Maze, J.~C. Adams, T.~Miller, N.~D.
  Kalka, A.~K. Jain, J.~A. Duncan, K.~Allen, et~al., Iarpa janus benchmark-b
  face dataset.

\bibitem{mtcnn}
K.~Zhang, Z.~Zhang, Z.~Li, Y.~Qiao, Joint face detection and alignment using
  multitask cascaded convolutional networks, IEEE Signal Processing Letters
  23~(10) (2016) 1499--1503.

\bibitem{facenet}
F.~Schroff, D.~Kalenichenko, J.~Philbin, Facenet: A unified embedding for face
  recognition and clustering, in: Proceedings of the IEEE conference on
  computer vision and pattern recognition, 2015, pp. 815--823.

\bibitem{heip1998indices}
C.~H. Heip, P.~M. Herman, K.~Soetaert, Indices of diversity and evenness,
  Oceanis 24~(4) (1998) 61--88.

\bibitem{rcnn}
S.~Ren, K.~He, R.~Girshick, J.~Sun, Faster r-cnn: towards real-time object
  detection with region proposal networks, IEEE Transactions on Pattern
  Analysis \& Machine Intelligence~(6) (2017) 1137--1149.

\bibitem{resnet}
K.~He, X.~Zhang, S.~Ren, J.~Sun, Deep residual learning for image recognition,
  in: Proceedings of the IEEE conference on computer vision and pattern
  recognition, 2016, pp. 770--778.

\bibitem{Simonyan:VGG16}
K.~Simonyan, A.~Zisserman, Very deep convolutional networks for large-scale
  image recognition, arXiv preprint arXiv:1409.1556 (2014).

\bibitem{GRU}
J.~Chung, C.~Gulcehre, K.~Cho, Y.~Bengio, Gated feedback recurrent neural
  networks, in: International Conference on Machine Learning, 2015, pp.
  2067--2075.

\bibitem{Robinson:FIW}
J.~P. Robinson, M.~Shao, Y.~Wu, Y.~Fu, Family in the wild (fiw): A large-scale
  kinship recognition database.

\bibitem{Wu:TIFS18}
X.~Wu, R.~He, Z.~Sun, T.~Tan, A light cnn for deep face representation with
  noisy labels, IEEE Transactions on Information Forensics and Security 13~(11)
  (2018) 2884--2896.

\bibitem{LSTM}
S.~Hochreiter, J.~Schmidhuber, Long short-term memory, Neural computation 9~(8)
  (1997) 1735--1780.

\bibitem{Yang:IJCAI18}
T.-Y. Yang, Y.-H. Huang, Y.-Y. Lin, P.-C. Hsiu, Y.-Y. Chuang, Ssr-net: a
  compact soft stagewise regression network for age estimation, in: Proceedings
  of the 27th International Joint Conference on Artificial Intelligence, AAAI
  Press, 2018, pp. 1078--1084.

\bibitem{role_of_attractiveness_masculinity}
R.~A. Hoss, J.~L. Ramsey, A.~M. Griffin, J.~H. Langlois, The role of facial
  attractiveness and facial masculinity/femininity in sex classification of
  faces, Perception 34~(12) (2005) 1459--1474.

\bibitem{ganin2016domain}
Y.~Ganin, E.~Ustinova, H.~Ajakan, P.~Germain, H.~Larochelle, F.~Laviolette,
  M.~Marchand, V.~Lempitsky, Domain-adversarial training of neural networks,
  The Journal of Machine Learning Research 17~(1) (2016) 2096--2030.

\bibitem{tzeng2015simultaneous}
E.~Tzeng, J.~Hoffman, T.~Darrell, K.~Saenko, Simultaneous deep transfer across
  domains and tasks, in: Proceedings of the IEEE International Conference on
  Computer Vision, 2015, pp. 4068--4076.

\bibitem{tzeng2017adversarial}
E.~Tzeng, J.~Hoffman, K.~Saenko, T.~Darrell, Adversarial discriminative domain
  adaptation, in: Proceedings of the IEEE Conference on Computer Vision and
  Pattern Recognition, 2017, pp. 7167--7176.

\bibitem{tsne}
L.~v.~d. Maaten, G.~Hinton, Visualizing data using t-sne, Journal of machine
  learning research 9~(Nov) (2008) 2579--2605.

\bibitem{gender_is_a_dimension}
J.~Y. Baudouin, G.~Tiberghien, Gender is a dimension of face recognition.,
  Journal of Experimental Psychology: Learning, Memory, and Cognition 28~(2)
  (2002) 362.

\bibitem{let_guess_age}
M.~C. Voelkle, N.~C. Ebner, U.~Lindenberger, M.~Riediger, Let me guess how old
  you are: Effects of age, gender, and facial expression on perceptions of
  age., Psychology and aging 27~(2) (2012) 265.

\end{thebibliography}





\end{document}